\documentclass{article}
\usepackage{xcolor}
\usepackage{PRIMEarxiv}
\usepackage[utf8]{inputenc} 
\usepackage[T1]{fontenc}    
\usepackage{hyperref}       
\usepackage{url}            
\usepackage{booktabs}       
\usepackage{amsmath, amssymb}
\usepackage{nicefrac}       
\usepackage{microtype}      
\usepackage{graphicx}       
\graphicspath{{media/}}     
\usepackage{fancyhdr}       
\usepackage{float}          
\usepackage{lipsum}         
\usepackage{booktabs} %
\usepackage{amssymb} %
\usepackage{colortbl}
\usepackage{tabularx} 
\usepackage{listings}
\usepackage[most]{tcolorbox} %
\tcbuselibrary{listings,breakable}
\lstdefinestyle{arxivcode}{
    basicstyle=\ttfamily\small,
    numbers=left,
    numberstyle=\tiny,
    numbersep=8pt,
    xleftmargin=2em,
    breaklines=true,
    columns=fullflexible,
    keepspaces=true,
    showstringspaces=false
}
\newtcblisting{myminted}[2][]{%
    listing engine=listings,
    listing options={style=arxivcode, language=#2},
    colback=white,       %
    colframe=black,      %
    boxrule=0.5pt,       %
    listing only,
    breakable,
    left=5mm,            %
    right=5mm,           %
    top=3mm,             %
    bottom=3mm,          %
    #1                   %
}

\usepackage{times}
\usepackage{latexsym}
\usepackage{amsmath}
\usepackage{amssymb}
\usepackage{pifont}

\usepackage{multirow}
\usepackage{booktabs, siunitx}
\usepackage{booktabs, siunitx} 
\usepackage[bottom]{footmisc}

\usepackage{xcolor}

\title{Text Distance from Nested and Hierarchical Repetitions: A Compression-Based Perspective}

 \author{
  \small
  \begin{minipage}[t]{0.95\textwidth}
  \centering
  \textbf{Xiaojun Hu$^{1,2,3,}$\thanks{These authors contributed equally.}\footnotemark[1], 
  Jing Wang$^{1,2,3,}$\footnotemark[1], 
  Jingwen Zhang$^{1,2}$, 
  Fengyao Zhai$^{1,2,3}$, 
  Xiao Xie$^{1,4}$, \\
  Hao Liao$^{5}$,
  Zengru Di$^{1,2}$, 
  Yu Liu$^{1,2,}$\thanks{Corresponding author: yu.ernest.liu@bnu.edu.cn}\footnotemark[2]} \\
  ~\\
  $^1$Department of Systems Science, Faculty of Arts and Sciences, Beijing Normal University, Zhuhai, China. \\
  $^2$International Academic Center of Complex Systems, Beijing Normal University, Zhuhai, China. \\
  $^3$School of Systems Science, Beijing Normal University, Beijing, China. \\
  $^4$School of Physics and Astronomy, Sun Yat-sen University, Zhuhai, China.\\
  $^5$College of Computer Science and Software Engineering, Shenzhen University,Shenzhen,China.
  \end{minipage}
  }

\begin{document}
\maketitle

\begin{abstract}
We present a new method for structural sequence analysis grounded in Algorithmic Information Theory (AIT). At its core is the Ladderpath approach, which extracts nested and hierarchical relationships among repeated substructures in linguistic sequences---an instantiation of AIT’s principle of describing data through minimal generative programs. These structures are then used to define three distance measures: a normalized compression distance (NCD), and two alternative distances derived directly from the Ladderpath representation. Integrated with a $k$-nearest neighbor classifier, these distances achieve strong and consistent performance across in-distribution, out-of-distribution (OOD), and few-shot text classification tasks. In particular, all three methods outperform both gzip-based NCD and BERT under OOD and low-resource settings. These results demonstrate that the structured representations captured by Ladderpath preserve intrinsic properties of sequences and provide a lightweight, interpretable, and training-free alternative for text modeling. This work highlights the potential of AIT-based approaches for structural and domain-agnostic sequence understanding.
\end{abstract}


\section{Introduction}

The rapid advancements in natural language processing (NLP) and machine learning have significantly improved the performance of text classification and regression tasks~\cite{info16020130, li2022survey}. A wide range of approaches, from traditional bag-of-words models to deep neural architectures and pre-trained language models such as BERT~\cite{devlin2019bert}, have been developed to tackle these tasks. However, deploying these models in low-resource or distributionally inconsistent scenarios remains a major challenge due to their heavy reliance on extensive annotated data and high computational demands~\cite{yang2021survey, mao2024low}. Although these models demonstrate good performance under ideal conditions, their generalization ability often diminishes in settings with scarce data or in the presence of domain shifts~\cite{cai2024multi}.

Traditional models and word embedding-based classifiers require feature engineering or fine-tuning \cite{shen2018baseline, pennington2014glove}, while large-scale models encode a highly compressive mapping of data in an ultra-high-dimensional space. From an information-theoretic perspective, the process of extracting patterns from data can be viewed as a form of compression—identifying and retaining only the most informative structures~\cite{deletang2024language, teahan2003using}. Inspired by this insight, Jiang et al. proposed a parameter-free classification method that combines a standard compressor (e.g., gzip) with a $k$-nearest neighbor ($k$-NN) classifier to approximate deep learning-like performance without training \cite{jiang2023low, wang2023adaptive}. This method uses the \textit{normalized compression distance} (NCD) to measure text similarity, offering a lightweight and generalized solution that is especially suited for low-resource and heterogeneous data. The theoretical foundation of this approach originates from the concept of information distance and Kolmogorov complexity~\cite{kolmogorov1963tables}. Bennett et al. introduced the notion of normalized information distance (NID), a universal similarity metric derived from Kolmogorov complexity \cite{681318}. However, due to the uncomputability of Kolmogorov complexity, NID cannot be directly applied in practice. To overcome this limitation, Li et al. proposed using compression algorithms to approximate complexity and introduced NCD, which serves as a computable alternative \cite{1362909}. Cilibrasi and Vitanyi later extended this idea to clustering tasks \cite{1412045}. By estimating the complexity of data objects through compression, NCD provides a model-free method for distance measurement, supporting training-free classification frameworks \cite{wang2020measurement}.

As mentioned above, compression-based techniques have demonstrated promising performance in text classification, for example through approaches that combine standard compressor with $k$-NN~\cite{jiang2023low}. The appeal of this method lies in its independence from extensive training or prior domain knowledge---it captures intrinsic regularities in data through general-purpose compression. However, general-purpose compressors like gzip are not optimized for semantic or hierarchical textual structures---features commonly found in natural human language---which can constrain their classification accuracy.

To address these limitations, we propose an alternative compression-based classification framework based on the Ladderpath approach \cite{liu2022ladderpath, liu2021exploring}, which falls under the broader framework of Algorithmic Information Theory (AIT). Ladderpath performs efficient compression by computing the minimal number of hierarchical reconstruction steps required to reproduce a given string or other data object \cite{zhang2024evolutionary}. This allows it to capture nested structural features more effectively than traditional compressors. Unlike pre-trained models or parameter-tuned systems, Ladderpath remains both model-free and parameter-free, which significantly enhances its adaptability in dynamic or data-sparse environments~\cite{keogh2004towards, mao2024low}. This makes it particularly suitable for real-world scenarios characterized by low data availability or inconsistent distributions.

The main contributions of this paper are summarized as follows: (1) We propose a new approach grounded in AIT, utilizing the Ladderpath approach to extract the nested and hierarchical relationships among repeated substructures in linguistic sequences and leverage them for compression. (2) We demonstrate that these structured relationships can be used both for compression-based distance computation---yielding a new normalized compression distance, $NCD_{lp}$---and for defining distances derived from the Ladderpath representation using ideas analogous to the Dice coefficient and Jaccard index, resulting in $L_{Dice}$ and $L_{Jaccard}$. (3) Experiments show that all three distance measures are effective for text classification tasks. $NCD_{lp}$ demonstrates performance comparable to the strongest previously reported compression-based approach, namely $NCD_{gzip}$, whereas $L_{Dice}$ and $L_{Jaccard}$ consistently achieve superior performance relative to $NCD_{lp}$. Notably, in out-of-distribution (OOD) and few-shot settings, all three methods outperform BERT. This not only provides a practical solution for scenarios with limited labeled data, but more importantly, highlights that the nested and hierarchical relationships extracted by Ladderpath capture intrinsic structural properties of sequences---enabling classification without any training.

\section{Methods}
\label{sec2}
\subsection{Recap Ladderpath approach: Capturing nested and hierarchical relationships}
\label{subsec1}
\begin{figure*}[http]  
\centering
\includegraphics[width=\linewidth]{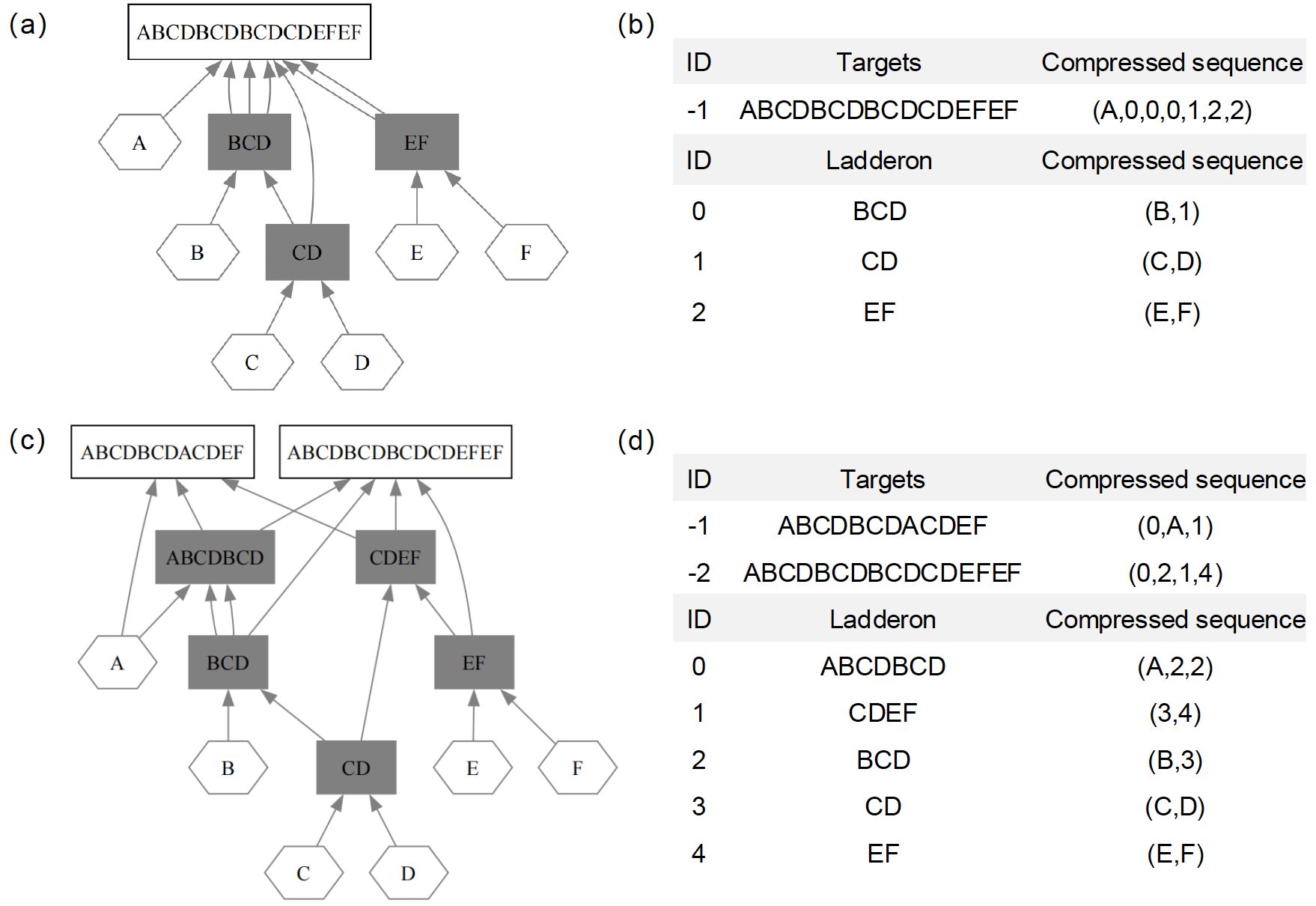}
    \caption{Illustration of nested and hierarchical relationships among repeated substructures in strings, as analyzed using the Ladderpath approach. Panels (a) and (c) display two example strings represented as named Laddergraphs. Panel (b) illustrates the compression process for the string in (a), while panel (d) shows the corresponding compression for the string in (c).}
    \label{fig:ex1}
\end{figure*}

The Ladderpath approach, which falls under the umbrella of AIT, seeks to find the shortest path for reconstructing an object (in this context, a string), with a key assumption that previously reconstructed substructures can be directly reused in subsequent steps---an idea that echoes Fran\c{c}ois Jacob's notion of evolutionary tinkering \cite{jacob1977evolution, Johnston2022PNAS, Zenil2019iScience}. It achieves this goal by identifying repeated substructures and the hierarchical relationships among these substructures. The length of this shortest path is defined as the \textit{ladderpath-index} $\lambda$. The hierarchical and nested relationships can be represented as a partially ordered multiset, or equivalently, a directed acyclic graph, referred to as the \textit{laddergraph} (see Fig. \ref{fig:ex1}a and \ref{fig:ex1}b for two examples). A detailed description of the Ladderpath approach can be found in~\cite{liu2022ladderpath, xu2024correlating, pephire2024}; only a brief recap is provided here.

As an illustrative example, consider the string `ABCDBCDBCDCDEFEF'. Its ladderpath is computed using an anonymized implementation (provided in the Supplementary Material) and can be represented as a partially ordered multiset: $\{$A, B, C, D, E, F // CD, EF // BCD(2)$\}$. The corresponding laddergraph is shown in Fig. \ref{fig:ex1}a. The ladderpath-index $\lambda$ for this string can also be computed, yielding a value of 10, indicating the minimum number of steps required to reconstruct the target string.

We can naturally apply the Ladderpath approach to compression because it computes the shortest reconstruction path by algorithmically identifying repeated substructures (referred to as \textit{ladderons}) and capturing their nested and hierarchical relationships. Each ladderon can be encoded in a dictionary with a unique ID, allowing us to simply reference its ID whenever it reappears (see the Section \ref{sec:lpcompressor} for a detailed description of the compression procedure).

\subsection{Ladderpath-based compressor}
\label{sec:lpcompressor}

Fig. \ref{fig:ex1}a illustrates the process of compressing a single string using the Ladderpath approach, with the previously discussed string `ABCDBCDBCDCDEFEF' as an example. After computing its ladderpath, each ladderon is assigned a unique ID: `BCD' receives an ID of 0, `CD' an ID of 1, and `EF' an ID of 2 (the higher the ID number, the lower the hierarchical level and typically the shorter the ladderon). Referring to Fig. \ref{fig:ex1}b, starting from the highest ID and moving downward, `EF' comprises the basic building blocks `E' and `F', thus represented directly as (E,F). Similarly, `CD' consists of the basic building blocks `C' and `D', denoted as (C,D). The ladderon `BCD' is composed of `B' and ladderon 1 (namely, `CD'), and is therefore represented as (B,1). This construction reduces the number of unique symbols: we no longer need to write the full sequence `B', `C', `D', effectively compressing one character.

Next, for the target string, we assign it a negative ID, here noted as $-1$, and represent it as (A,0,0,0,1,2,2). Every time ID 0 appears, we avoid rewriting `BCD', thereby saving two characters per occurrence. Since ID 0 occurs three times, a total of six characters is saved. Similarly, each occurrence of ID 1 saves one character, thus two occurrences save two characters, and so forth. This strategy of eliminating redundant rewrites is the basis of compression in the Ladderpath approach.

Given the above definitions and constructions, all relevant information can be encoded into a single sequence. The resulting compression under the Ladderpath framework can be expressed as
\begin{equation*}
    z = \text{(1; A,0,0,0,1,2,2; B,1; C,D; E,F)}
\end{equation*}
where the first number indicates the total number of target strings---in this case, 1. The first semicolon-separated section encodes the target string using ladderon IDs. The following sections, also separated by semicolons, define all the ladderons: `B,1' corresponds to ladderon ID 0, `C,D' to ID 1, and `E,F' to ID 2.

Excluding the initial number 1, the total length of this compressed string $z$ is 13, which equals $\lambda$ plus the total number of ladderons. In this example, $\lambda$ equals 10, and there are a total of 3 ladderons. This can be clearly demonstrated since $\lambda$ itself represents the shortest number of steps required to reconstruct the target string from basic units. The compressed string directly reflects this minimal reconstruction path. Note that the number of ladderons (3 in this case) is added because combining $n$ ladderons involves only $(n-1)$ steps; for example, forming `EF' from `E' and `F' is counted as one step, but in the compressed sequence we must explicitly write two characters, `E' and `F'. Similarly, constructing `BCD' from `B' and ladderon 1 is one step, but we must write both `B' and 1 in the compressed output. Finally, Fig. \ref{fig:ex1}b demonstrates the compression of two target strings using the same approach.

In principle, the sequence $z$ can be further compressed into a binary sequence, for example, by using Huffman coding or converting the final compressed sequence into another format (see Appendix \ref{App:compressor} for more details). However, such additional steps are not considered here, as the focus is on defining a distance measure and performing text classification based on Ladderpath compression. In summary, since the ladderpath-index $\lambda$ is defined as the length of the shortest reconstruction path derived from the hierarchical and nested relationships among repeated substrings, $\lambda$ can be used as an effective proxy for the optimally compressed length.

Finally, according to \cite{1412045}, to define NCD using a compressor, the compressor should satisfy the following four key properties: \textit{Idempotency}, ensuring that duplicate data does not affect compression efficiency; \textit{Monotonicity}, requiring that compressing multiple strings does not yield a smaller result than compressing a single string; \textit{Symmetry}, indicating that compression results should be independent of the order of input data; and \textit{Distributivity}, ensuring consistent processing of different string combinations regardless of data structure, order, or chunking method. A compressor that meets these properties within an acceptable error margin is considered a \textit{normal compressor} \cite{1412045}. Not all compressors strictly satisfy these four properties. For instance, widely used compressors such as $Zstandard$ exhibit deviations in distributivity. We have conducted systematic experiments and found that the Ladderpath-based compressor satisfies these properties within an acceptable error margin (see Appendix \ref{App:normal} for details).\\

\noindent \textbullet{~~{Normalized Compression Distance (NCD): Ladderpath-based}}

Once the Ladderpath-based compressor is shown to function as a normal compressor, it can be used to define a Ladderpath-based NCD. When using a compressor, the NCD between two strings, $X$ and $Y$, is defined as follows \cite{1362909, 1412045}: 
\begin{equation*}
    NCD_c(X,Y) = \frac{c(X,Y) - \min \big[ c(X), c(Y) \big]}{\max \big[ c(X), c(Y) \big]} 
\end{equation*}
where $c(X)$ is the length of $X$ after it is compressed using the compressor $c$. By substituting in the Ladderpath-based compressor, we obtain
\begin{equation*}
    NCD_{lp}(X,Y) = 
    \frac{\lambda'(X,Y) - \min \big[ \lambda'(X), \lambda'(Y) \big]}
    {\max \big[ \lambda'(X), \lambda'(Y) \big]}
\end{equation*}
where $\lambda'(X) \equiv \lambda(X) - 1$ and $\lambda'(X,Y) \equiv \lambda(X,Y) - 2$, and $\lambda$ is the ladderpath-index. This small adjustment---subtracting $1$ and $2$---is applied to balance the operation of taking out the target string(s) during reconstruction. If $n$ targets are taking out, then $n$ should be subtracted. We now apply $NCD_{lp}$ to a text classification task, with results presented in Section \ref{sec:results}.

\subsection{Define Ladderpath-distance L}

Before conducting the text classification task, two alternative distance measures can be defined directly from the hierarchical and nested relationships among repeated substrings (i.e., ladderons). One of them is based on the idea behind the \textit{Dice coefficient}. The similarity between two sets, $P$ and $Q$, is defined as the ratio of the size of their intersection to the average size of the two sets, namely $|P \cap Q|/((|P|+|Q|)/2)$, and consequently, the distance is defined as one minus this similarity. By applying the inclusion-exclusion principle, this distance can be derived as:
\begin{equation}
\label{eq:dice}
1- \frac{|P \cap Q|}{\frac{|P|+|Q|}{2}} = \frac{\frac{|P|+|Q|}{2} - |P \cap Q|}{\frac{|P|+|Q|}{2}} = \frac{\frac{|P|+|Q|}{2} - (|P|+|Q|-|P \cup Q|)}{\frac{|P|+|Q|}{2}} = \frac{|P \cup Q|- \frac{|P|+|Q|}{2}}{\frac{|P|+|Q|}{2}}
\end{equation}

Thus, a \textit{Ladderpath-distance} based on the Dice coefficient can be defined as:
\begin{equation*}
    L_{Dice}(X, Y) = \frac{\lambda'(X,Y) - \frac{\lambda'(X) + \lambda'(Y)}{2}}{\frac{\lambda'(X) + \lambda'(Y)}{2}}
\end{equation*}
where $\lambda'(X)$ represents the shortest path length required to reconstruct string $X$ individually, corresponding to the set size $|P|$ in Eq. \eqref{eq:dice}, and similarly for $\lambda'(Y)$. Furthermore, $\lambda'(X,Y)$ represents the shortest path length when reconstructing strings $X$ and $Y$ jointly, corresponding to $|P \cup Q|$ in Eq. \eqref{eq:dice}.

Alternatively, the other distance measure is based on the idea behind the \textit{Jaccard index}, in which the similarity between two sets $P$ and $Q$ is defined as $|P \cap Q| / |P \cup Q|$. The only difference from the Dice coefficient lies in the denominator: the union size rather than the average size. Following a derivation similar to the one above, a \textit{Ladderpath-distance} based on the Jaccard index can be defined as:
\begin{equation*}
    L_{Jaccard}(X, Y) = \frac{\lambda'(X,Y) - \frac{\lambda'(X) + \lambda'(Y)}{2}}{\frac{\lambda'(X,Y)}{2}}
\end{equation*}
Note that both distance measures ensure that when $X=Y$, the distance is $0$, and when $X$ and $Y$ share no common substructure, the distance is $1$.

It is worth noting that the similarity or distance measures based on the Jaccard index and the Dice coefficient do exhibit some differences (although they can be transformed into one another). It is well known that the Jaccard-based distance satisfies the triangle inequality, making it a true metric. In contrast, the Dice-based distance does not satisfy the triangle inequality, and is therefore considered a semimetric version of the Jaccard distance. Nevertheless, both measures are commonly used in practice. In some cases, the Dice-based distance is even preferred because empirical evidence suggests that, compared to the Jaccard index, the Dice coefficient tends to yield higher similarity scores, especially in high-dimensional, sparse data---such as in bag-of-words models or image segmentation masks. For example, in medical image segmentation (CT, MRI), the Dice coefficient is widely used as an evaluation metric. Similarly, certain $n$-gram–based tasks in natural language processing employ the Dice coefficient~\cite{dice1945measures, milletari2016v}. In deep learning applications involving medical imaging---such as Mask R-CNN and other segmentation tasks---Dice loss is extensively used as a loss function (while Jaccard loss is less common), due to its smoother gradient properties, which help with network convergence \cite{li2019dice}. For the sake of completeness, we employ both distance measures in this work to perform text classification tasks.

\section{Experiments and results}
\label{sec:results}

In the work by Jiang et al. \cite{jiang2023low}, the authors employed $NCD_{gzip}$ (a gzip-based normalized compression distance), combined with a $k$-NN classifier, to perform text classification across three distinct scenarios: in-distribution datasets, OOD datasets, and few-shot learning settings. They demonstrated that this simple, training-free approach can achieve performance comparable to---or even surpass---that of large language models such as BERT, despite BERT being a substantially more complex model.

In this study, we employ the newly defined distances---$NCD_{lp}$ (Ladderpath-based normalized compression distance), $L_{Dice}$ and $L_{Jaccard}$---to carry out classification tasks and compare their performance against the original $NCD_{gzip}$. The results are presented in the following subsections.

\subsection{Experimental setup}

All datasets used in this study for text classification are publicly available from their original sources. A brief description of each dataset is as follows. AGNews comprises over one million news articles from more than 2,000 sources collected via \textit{ComeToMyHead}. DBpedia aggregates structured information from all Wikipedia language editions, Wikidata, Wikimedia Commons, and related projects, and is accessible in \textit{TorchText} . R8 and R52 are subsets of the Reuters-21578 corpus and can be obtained from the \textit{Text Categorization Corpora}. KinyarwandaNews and KirundiNews consist of news articles from Rwandan and Burundian websites and newspapers and are freely downloadable from the \textit{corresponding repository}. SwahiliNews was specifically created for text classification in the African language Swahili, with each item categorized into six topics, and is freely available through \textit{HuggingFace}. DengueFilipino is a benchmark dataset for low-resource multiclass classification, primarily in Filipino but containing some English loanwords common in colloquial usage, and is likewise accessible on \textit{HuggingFace}. Finally, SogouNews is a large-scale Chinese corpus provided by Sogou Inc., comprising extensive text from online news outlets, forums, and blogs, and is publicly distributed via \textit{TorchText}. Detailed download links for all datasets are provided in Appendix \ref{App:datasets}.

To ensure high data quality and the reliability of experimental results, we implemented a rigorous data cleaning and screening procedure. First, we removed duplicate records to retain only unique instances, as duplication can lead the $k$-NN classifier to produce spuriously high performance. Next, we eliminated contaminated entries---cases in which the same text was associated with inconsistent labels---which would otherwise introduce noise and increase the risk of misclassification. Furthermore, for the large-scale DBpedia and SogouNews datasets, we selected approximately 10\% of the available records for our experiments (see Appendix \ref{App:preprocess} for details). This sampling strategy preserved the representativeness of the data while substantially reducing computational time and cost.

\subsection{For in-distribution datasets}
Now, we conduct classification experiments using the newly defined distances, and the results are presented in Table \ref{tab:in_distribution}. For comparison, we adopt the accuracy values for TextCNN, LSTM, W2V, and BERT from the work by Jiang et al. \cite{jiang2023low}, and use TextLength (i.e., classifying based on text length) as the baseline. It is important to note that the accuracy of $NCD_{gzip}$ reported in \cite{jiang2023low} was slightly inflated \cite{opitz2023gzip}, due to the use of an uncleaned dataset and an optimistic tie-breaking strategy in the $k$-NN classifier (see Appendix \ref{App:kNNoptimistic} for details). In fact, the authors later acknowledged this issue. Consequently, we recomputed the classification accuracy for $NCD_{gzip}$ using the same methodology but on the cleaned dataset.



\begin{table}[!ht]
\centering
\renewcommand{\arraystretch}{1.2} 
\scalebox{1.0}{
\begin{tabular}{lcccccc}
    \toprule
    \textbf{Model}          & \textbf{Pre-training} & \textbf{Training} & \textbf{AGNews} & \textbf{DBpedia}  & \textbf{R8} & \textbf{R52} \\ 
    \midrule

    TextLength             & - & - & 0.275 & 0.093 & 0.455 & 0.362 \\ 
    
    TextCNN                & - & \ding{51} & 0.817 & 0.981 & 0.951 & 0.895 \\ 
    LSTM                   & - & \ding{51} & 0.861 & 0.985 & 0.937 & 0.855 \\ 
    W2V                    & \ding{51} & - & 0.892 & 0.961 & 0.930 & 0.856 \\ 
    BERT                   & \ding{51} & \ding{51} & 0.944 & 0.992 & 0.982 & 0.960 \\ 

    Bag-of-Words                & - & \ding{51} & - & -  & 0.930 & 0.867 \\
    
    \midrule
    
    \textbf{$NCD_{gzip}$}  & - & - & {0.876} & {0.831} & {0.916} & {0.834} \\ 
    
    \textbf{$NCD_{lp}$}    & - & - &
    {0.876}  &
    {0.838} & 
    {0.901} &
    {0.863} \\ 
    
    \textbf{$L_{Dice}$}           & - & - & 
    \textbf{0.885}  & 
    \textbf{0.857} & 
    \textbf{0.918} & 
    \textbf{0.864} \\ 
    
    \textbf{$L_{Jaccard}$}           & - & - & 
    \textbf{0.885}  & 
    \textbf{0.857} & 
    \textbf{0.918} & 
    \textbf{0.864} \\ 
    \bottomrule
\end{tabular}
}
\vspace{0.3cm}
\caption{Comparison results on text classification datasets.}
\label{tab:in_distribution}
\end{table}

As shown in Table \ref{tab:in_distribution}, we report the results for the three newly defined distance measures: $NCD_{lp}$, $L_{Dice}$ and $L_{Jaccard}$. Among methods that do not rely on pre-training or task-specific training, $L_{Dice}$ and $L_{Jaccard}$ attain the highest classification accuracies; $NCD_{lp}$ outperforms $NCD_{gzip}$ in most cases. These results indicate that the proposed measures exhibit competitive classification performance, suggesting that the hierarchical and nested structures captured by Ladderpath approach can provide meaningful similarity representations for text classification tasks. Note that in our experiments, we set the parameter $k=7$ for all $k$-NN-based methods, rather than using $k=2$ as in the study by Jiang et al., since $k=7$ consistently yields near-optimal performance across these methods. For completeness, however, we also report the results with $k=2$ in Table \ref{tab:table1k_2}, where $L_{Dice}$ and $L_{Jaccard}$ likewise achieve the best performance.

\begin{table}[!ht]
\centering
\renewcommand{\arraystretch}{1.2} 
\scalebox{1.0}{
\begin{tabular}{lcccccc}
    \toprule
    \textbf{Model} & \textbf{Pre-training} & \textbf{Training} & \textbf{AGNews} & \textbf{DBpedia} & \textbf{R8} & \textbf{R52} \\ 
    \midrule
    $NCD_{gzip}$ & - & - & {0.834} & {0.825} & {0.896} & {0.876} \\ 
    $NCD_{lp}$ & - & - & {0.863} & {0.829} & {0.887} & {0.872} \\ 
    $L_{Dice}$ & - & - & {0.857} & \textbf{0.853} & \textbf{0.907} & \textbf{0.881} \\ 
    $L_{Jaccard}$ & - & - & \textbf{0.864} & \textbf{0.853} & \textbf{0.907} & \textbf{0.881} \\ 
    \bottomrule
\end{tabular}
}
\vspace{0.3cm}
\caption{Classification results under the same setting as Table \ref{tab:in_distribution}, but with $k=2$, following the settings in Jiang et al. Some of the results reported in their study appear higher than ours because, as explained in Appendix \ref{App:kNNoptimistic}, their $k$-NN classification systematically adopted the most optimistic scenario. Jiang et al. have since acknowledged this issue and updated certain results on their GitHub repository.}
\label{tab:table1k_2}
\end{table}

In Table \ref{tab:in_distribution}, we also presented the results of the \textit{bag-of-words} approach, as described in \cite{opitz2023gzip}. This method represents a conventional text classification technique that relies on word occurrence patterns rather than deep contextual understanding or direct distance-based comparison. Specifically, it follows a three-step preprocessing pipeline: (1) removing punctuation and replacing it with spaces; (2) filtering out words below a certain frequency threshold; and (3) converting all uppercase letters to lowercase. After this transformation, texts are represented as independent word vectors, which are then used to compute similarities and perform classification. Notably, we observe that this approach also yields strong performance in classification tasks. However, it is important to emphasize that its success primarily stems from an inherent form of semantic segmentation, in which words naturally serve as meaningful semantic units. In contrast, compression-based and Ladderpath-based methods operate without incorporating any prior semantic knowledge.

Note that in Table \ref{tab:in_distribution} and \ref{tab:table1k_2}, we observe nearly identical classification accuracies for $L_{Dice}$ and $L_{Jaccard}$, which may appear somewhat unexpected. Nevertheless, this outcome can be clarified through a closer examination of their empirical behavior. Specifically, we compared the distance matrices between one focal text and ten reference texts within the AGNews dataset, as presented in Table \ref{tab:l_matrix}. The analysis reveals that, for the same text pairs, the distance values produced by $L_{Dice}$ and $L_{Jaccard}$ are numerically very close, though not strictly identical. These marginal discrepancies exert only a negligible effect on the performance of the $k$-NN classifier. Consequently, despite their theoretical distinctions, $L_{Dice}$ and $L_{Jaccard}$ yield highly consistent assessments of text similarity in practice, which accounts for the near-equivalence observed in their classification accuracies.
\begin{table}[!ht]
\centering
\renewcommand{\arraystretch}{1.2} 
\scalebox{1.0}{
\begin{tabular}{lccccccccc}
    \toprule
    Text & {1} & 2 & 3 & 4 & 5 & {6} & {7} & {8} & {9} \\ 
    \midrule
    $L_{{Dice}}$ & 0.820 & 0.927 & 0.934 & 0.944 & 0.936 & 0.943 & 0.875 & 0.912 & 0.913 \\
    $L_{Jaccard}$ & 0.695 & 0.865 & 0.876 & 0.895 & 0.879 & 0.891 & 0.779 & 0.839 & 0.840 \\ 
    \bottomrule
\end{tabular}
}
\vspace{0.3cm}
\caption{Comparison of pairwise distances between a focal training text and multiple test texts in the AGNews dataset, evaluated using $L_{Dice}$ and $L_{Jaccard}$.}
\label{tab:l_matrix}
\end{table}

\begin{figure}[htbp]
  \centering
  \includegraphics[width=0.8\linewidth]{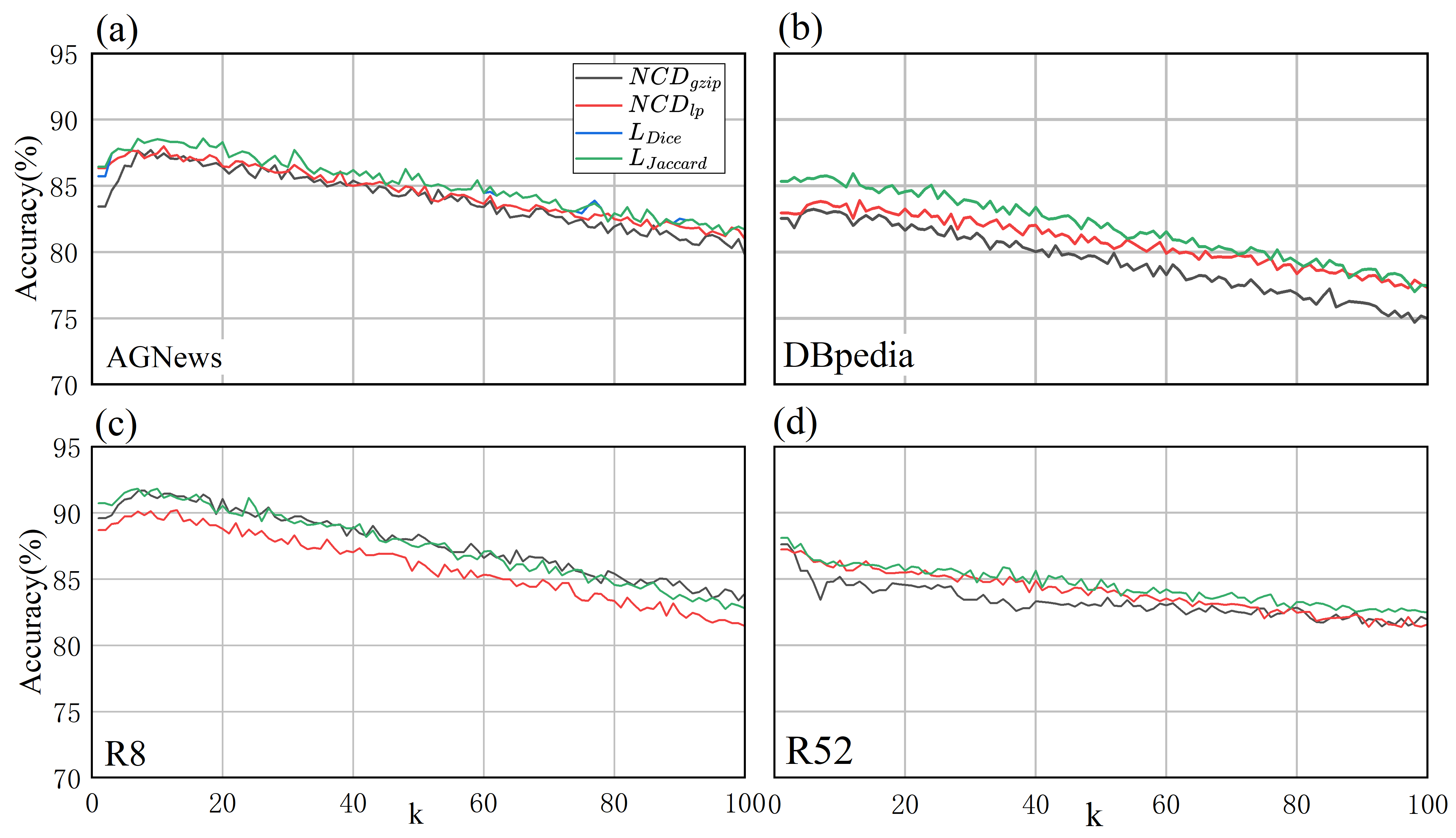}
  \caption{Accuracy comparison of $k$-NN-based methods under varying values of $k$ across different datasets: (a) AGNews dataset; (b) DBpedia dataset; (c) R8 dataset; (d) R52 dataset. In (a), a small portion of the blue $L_{Dice}$ line is visible, while the remainder overlaps entirely with the green $L_{Jaccard}$ line. In (b), (c), and (d), the blue $L_{Dice}$ line is completely obscured due to the overlap with the green line. The reason for this behavior has been discussed above.}
  \label{fig:k}
\end{figure}

Finally, we report the classification accuracies obtained for different values of $k$ in Fig. \ref{fig:k}. The results reveal that (1) the Ladderpath-compression-based distance $NCD_{lp}$ surpasses $NCD_{gzip}$ on DBpedia and R52, performs comparably on AGNews, and is outperformed by $NCD_{gzip}$ only on R8; (2) the Ladderpath-distance $L_{Dice}$ and $L_{Jaccard}$ achieve the best performance across all four datasets---except that on R8, $NCD_{gzip}$ remains similarly strong.

\subsection{For out-of-distribution (OOD) datasets}

OOD robustness is a major challenge for modern machine learning systems, which must maintain reliable performance when test data differ substantially from the training distribution. Compression-based and Ladderpath-based methods are particularly well suited to this setting because they require no pre-training or language-specific fine-tuning.

We evaluated OOD robustness on five datasets in different natural languages—Kinyarwanda, Kirundi, Filipino, Swahili, and Chinese (SogouNews)—as listed in Table \ref{tab:comparison-ood}. BERT serves as a baseline, with its accuracy taken from Jiang et al. \cite{jiang2023low}. Among purely distance-based approaches, $L_{Dice}$ and $L_{Jaccard}$ consistently outperform $NCD_{gzip}$, while $NCD_{lp}$ and $NCD_{gzip}$ exhibit comparable accuracy, each surpassing the other in roughly half of the tasks. It should be noted that, for reasons discussed earlier, the accuracy values of $NCD_{gzip}$ originally reported by Jiang et al. \cite{jiang2023low} were slightly inflated \cite{opitz2023gzip}. The authors have acknowledged this issue and released corrected results in subsequent GitHub comments (link is provided in Appendix \ref{App:datasets}), which we adopt for a fair comparison.

\begin{table}[!ht]
\centering
\renewcommand{\arraystretch}{1.2} 
\scalebox{1.0}{
\begin{tabular}{lccccc}
    \toprule
    \textbf{Model} & \textbf{Kinyar-wanda-News} & \textbf{Kirundi-News} & \textbf{Dengue-Filipino} & \textbf{Swahili-News} & \textbf{SogouNews} \\
    \midrule
    BERT & 0.838 & 0.879 & 0.979 & 0.897 & 0.952 \\
    \midrule
    $NCD_{gzip}$ & 0.830 & 0.867 & 0.963 & 0.887 & 0.943 \\
    $NCD_{lp}$ & 0.820 & 0.873 & 0.979 & 0.863 & 0.943 \\
    $L_{Dice}$ & \textbf{0.842} & \textbf{0.905} & \textbf{0.981} & \textbf{0.892} & \textbf{0.959} \\
    $L_{Jaccard}$ & \textbf{0.842} & \textbf{0.905} & \textbf{0.981} & \textbf{0.892} & \textbf{0.959} \\
    \bottomrule
\end{tabular}
}
\vspace{0.3cm}
\caption{Comparison of text classification accuracy across various methods and OOD datasets. Among the four distance-based methods below, the highest accuracy in each row is highlighted in bold. If BERT's accuracy exceeds the best of the four distance-based methods, it is additionally marked with an underline. All $k$-NN-based methods use $k=7$, in contrast to Jiang et al.'s study, which used $k=2$. For reference, the results with $k=2$ are provided in Appendix \ref{App:table2k_2}.}
\label{tab:comparison-ood}
\end{table}

An important implication of these results is the cross-lingual generality of the Ladderpath approach. Despite having no prior linguistic knowledge of these diverse languages, the Ladderpath-derived distances capture structural regularities that align closely with semantic similarity. In other words, the hierarchical and nested patterns detected by Ladderpath provide a language-agnostic metric of textual relatedness. This explains why $L_{Dice}$ and $L_{Jaccard}$ achieve strong and stable performance across typologically different languages without any additional adaptation.

\subsection{Few-shot setting}

Few-shot classification tasks pose a significant challenge for conventional machine learning models, particularly when the number of labeled examples per class is extremely limited or when the label distribution is highly imbalanced~\cite{liu2025guiding}. While many recent few-shot methods rely on large pre-trained models or meta-learning frameworks, we explore a fundamentally different direction based on compression-based and training-free approaches. These methods, including $NCD_{lp}$, $L_{Dice}$, and $L_{Jaccard}$, are naturally suited to low-resource and OOD scenarios.

To evaluate the effectiveness of different compression-based distance metrics in such low-resource conditions, we conduct experiments on five OOD datasets using the 5-shot setting. The results, presented in Table \ref{tab:fewshot-ood}, show that: (1) the traditional compression-based method $NCD_{gzip}$ outperforms BERT on 3 out of the 5 datasets and performs comparably to the Ladderpath-based compression method $NCD_{lp}$; (2) the distance measures computed directly from Ladderpath, namely $L_{Dice}$ and $L_{Jaccard}$, significantly outperform both BERT and $NCD_{gzip}$. These findings suggest that pre-training-free methods---especially those based on our Ladderpath approach, which captures the nested and hierarchical relationships among repeated substructures---are better equipped to extract discriminative features from limited samples.

\begin{table}[!ht]
\centering
\renewcommand{\arraystretch}{1.2} 
\scalebox{1.0}{
\begin{tabular}{lccccc}
    \toprule
    \textbf{Model} & \textbf{Kinyar-wanda-News} & \textbf{Kirundi-News} & \textbf{Dengue-Filipino} & \textbf{Swahili-News} & \textbf{SogouNews} \\ 
    \midrule 
    BERT & 0.240 & 0.386 & 0.409 & 0.396 & 0.221 \\ 
    \midrule 
    $NCD_{gzip}$ & 0.285 & 0.329 & 0.362 & 0.565 & 0.406 \\ 
    $NCD_{lp}$ & 0.242 & 0.320 & 0.369 & 0.559 & 0.414 \\ 
    $L_{Dice}$ & \textbf{0.299} & \textbf{0.417} & \textbf{0.418} & \textbf{0.571} & \textbf{0.481} \\ 
    $L_{Jaccard}$ & \textbf{0.299} & \textbf{0.417} & \textbf{0.418} & \textbf{0.571} & \textbf{0.481} \\
    \bottomrule 
\end{tabular}
}
\vspace{0.3cm}
\caption{Comparison of text classification accuracy across various methods on OOD datasets under the 5-shot setting. Among the four distance-based methods below, the highest accuracy in each row is highlighted in bold. All $k$-NN-based methods use $k=7$, in contrast to Jiang et al.'s work, which used $k=2$. For reference, results with $k=2$ are provided in Appendix \ref{App:table3k_2}.}
\label{tab:fewshot-ood}
\end{table}

\begin{figure}[http]
  \centering
  \includegraphics[width=1.0\linewidth]{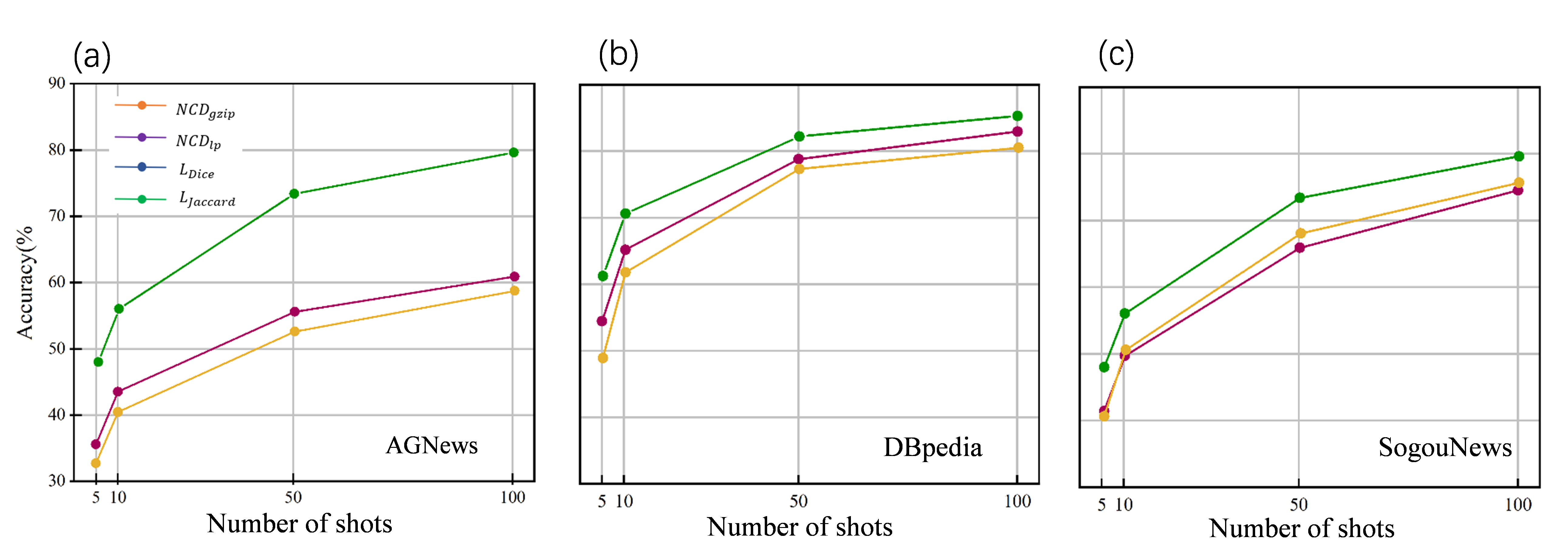}
  \caption{Few-shot performance comparison across multiple datasets. As with Fig. \ref{fig:k}, the reason the blue $L_{Dice}$ line is not visible is that it is completely obscured by the overlapping green $L_{Jaccard}$ line.}
  \label{Fig:three_data_0-100shot}
\end{figure}

Beyond OOD tasks, few-shot learning is also applicable to in-distribution scenarios, particularly when aiming for lightweight models or when annotation costs are high. To further evaluate the scalability and robustness of the proposed distance measures, we carried out additional experiments under different $n$-shot settings ($n \in \{5, 10, 50, 100\}$) on three datasets: AGNews, DBpedia, and SogouNews. The results, shown in Fig. \ref{Fig:three_data_0-100shot}, indicate that the Ladderpath-based distances $L_{Dice}$ and $L_{Jaccard}$ markedly outperform all other methods across all tested cases. These results complement the OOD evaluation in Table \ref{tab:fewshot-ood} by illustrating performance trends as the number of labeled examples increases. We selected these three datasets because their data scales are sufficiently large to support 100-shot experiments, and they differ in average text length and language. This diversity enables a more comprehensive evaluation across heterogeneous input conditions while ensuring consistency across varying few-shot levels.

Taken together, the few-shot results show that Ladderpath-derived distances are particularly effective under scarce supervision, consistently outperforming both neural and traditional compression-based baselines across OOD and in-distribution settings. This advantage remains stable as the number of labeled examples increases, indicating strong scalability in low-resource regimes. More importantly, these findings suggest that explicitly modeling hierarchical reuse yields representations that are inherently sample-efficient: rather than relying on parameter learning, this approach exploits structural regularities in the data itself, making it well suited for few-shot and low-resource sequence understanding.

\section{Discussion}

The effectiveness of text classification fundamentally depends on both the quality of feature extraction and the suitability of the classification model. Deep learning architectures---such as convolutional neural networks and pre-trained language models like BERT---have achieved remarkable success by capturing deep semantic patterns in text. However, these methods typically require large-scale annotated datasets, substantial computational resources, and often struggle with generalization under domain shift or low-resource conditions.

In contrast, the Ladderpath approach offers a lightweight, training-free alternative grounded in AIT. By directly identifying repeated substructures and capturing their nested and hierarchical relationships, Ladderpath constructs a structural representation that is both compact and informative. These relationships are then leveraged to define distance measures---either through compression-based comparison (yielding $NCD_{lp}$) or via structure-inspired metrics ($L_{Dice}$, $L_{Jaccard}$)---and combined with a simple $k$-NN classifier. Our empirical results demonstrate that these Ladderpath-based distances consistently achieve strong classification performance, often comparable to deep learning baselines. In particular, $L_{Dice}$ and $L_{Jaccard}$ outperform traditional gzip-based NCD in both OOD and few-shot settings, highlighting the expressiveness and robustness of the hierarchical structure captured by Ladderpath.

From a methodological perspective, the strength of the Ladderpath approach lies in the nature of its inductive bias. Rather than learning representations through gradient-based optimization, it enforces a structural bias toward reuse and minimal generative description. This bias favors representations that are inherently sample-efficient, as they are derived directly from the internal organization of the data rather than from statistical estimation over large corpora. As a result, Ladderpath-based distances remain stable under limited supervision and distribution shift, where learned representations often struggle.

This distinction is further reflected in our comparison with bag-of-words-based approaches, which typically involve aggressive preprocessing steps such as lowercasing, punctuation removal, and filtering of short or infrequent words. These procedures, while not explicitly semantic, introduce strong linguistic priors that help segment the input into semantically meaningful units. In contrast, Ladderpath retains the raw sequential order of the text while structurally compressing it based on internal repetition and recursive reuse. As such, it captures meaningful regularities in a domain-agnostic and language-independent manner---providing a structural and interpretable perspective on text compression and similarity.

Taken together, these observations position Ladderpath as a complementary alternative to model-centric approaches in text classification. By focusing on structural compression rather than parameter learning, it highlights a different axis of generalization---one grounded in reusable structure rather than learned semantics---which may be particularly valuable in transparent, low-resource, or cross-lingual settings.

\section{Conclusion}

In this work, we leveraged the Ladderpath approach as a structural representation grounded in AIT to define a family of distance measures for text classification. By operating directly on Ladderpath-derived structures rather than learned parameters, these distances achieve strong and stable performance across in-distribution, out-of-distribution, and few-shot settings, demonstrating their effectiveness for sequence comparison when labeled data are limited.

Beyond classification, the Ladderpath approach provides a general perspective on sequence modeling that emphasizes minimal generative structure rather than parameter learning. By preserving nested and hierarchical relationships among repeated substructures, it supports structure-aware similarity estimation and may serve as a preprocessing or tokenization module for downstream models. More broadly, the core principles underlying Ladderpath---recursive structure reuse and information-theoretic minimality---resonate with a growing line of work that views compression as a key organizing principle for representation learning and intelligence~\cite{deletang2024language, LiMing2025, zenil2019causal}. Together, these results highlight the value of structurally grounded, compression-based representations for robust sequence understanding under data scarcity and distribution shift.

\section{Acknowledgment}
We gratefully acknowledge Prof. Fan Jin and Prof. Ziwei Dai for their insightful suggestions and constructive discussions. This study was funded by the National Natural Science Foundation of China (Grant
No. 12205012 to Y.L.) and Basic and Applied Basic Research Foundation of Guangdong Province (Grant No. 2025A1515012923 to Y.L.).


\section{Declaration of competing interest}
The authors declare that they have no known competing financial interests or personal relationships that could have appeared to influence the work reported in this paper.

\section{Data availability}
The source code for Ladderpath calculations is openly accessible at \url{https://github.com/yuernestliu/lppack}


\newpage
\appendix
\addcontentsline{toc}{section}{Appendix}
\vspace{1em}
\section*{Appendix}

\section{Implementation of Ladderpath-based compressor}
\label{App:compressor}

\begin{figure*}[http]
\centering
\includegraphics[width=0.9\linewidth]{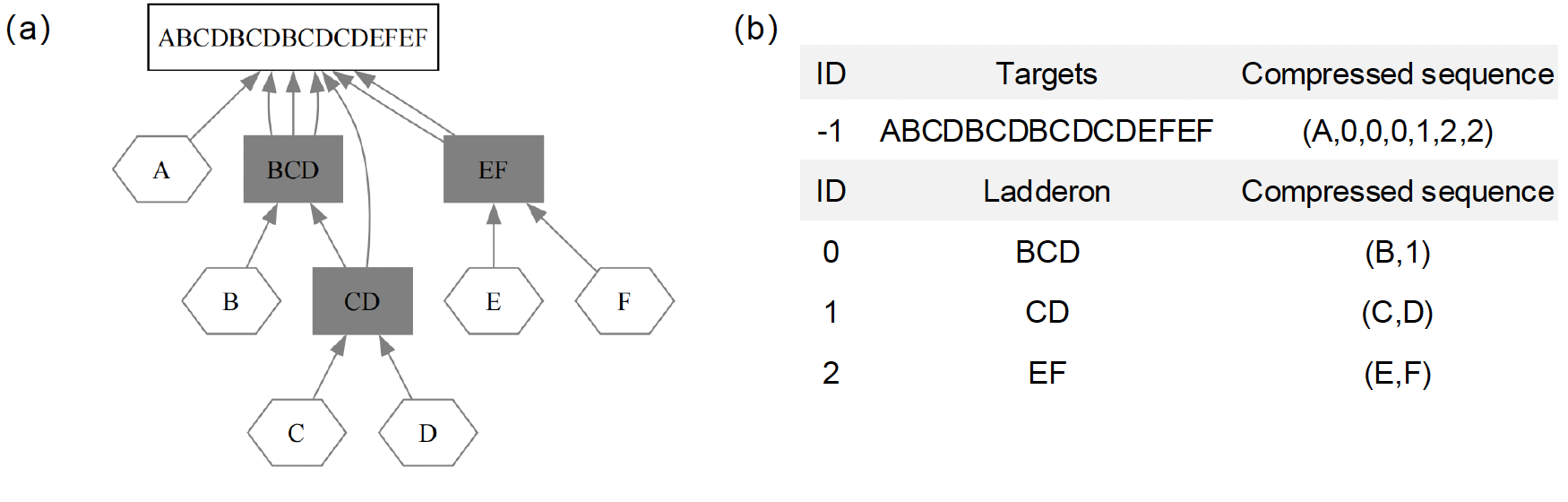}
    \caption{Illustration of nested and hierarchical relationships among repeated substructures in strings, as analyzed using the Ladderpath approach. }
    \label{fig:SI1}
\end{figure*}

\vspace{0.2cm}
Still taking the string “ABCDBCDBCDCDEFEF” as an example (Fig. \ref{fig:ex1}a in the main text), after processing with the Ladderpath approach, we obtain three ladderons as Fig. \ref{fig:SI1} shows. The original string is essentially composed of basic buildings and these ladderons. 

At the operational level, we can compute it using the code we provide as follows:
\begin{myminted}{Python}
import ladderpath as lp
import ladderpath_tools.compress as lp_c

X = ['ABCDBCDBCDCDEFEF']
lpjson = lp.get_ladderpath(X)

compressed_list = lp_c.compress(lpjson, SEP='@')

# The result is:
# compressed_list = [1, '@', 'A', 2, 2, 2, 1, 0, 0, '@', 'EF', 
#                   '@', 'CD', '@', 'B', 1]
\end{myminted}

We can see that the result is not exactly the same as the one shown in the original text, $z = (1; A,0,0,0,1,2,2; B,1;\allowbreak  C,D; E,F)$,but they correspond one-to-one. Let's now explain the structure of the above `compressed\_list': `@' serves as a delimiter to separate different sequences and ladderons. The first element, `1', indicates that the target sequence contains only one string. The subsequent elements describe the reconstruction process of the original string, where ```A', 2, 2, 2, 1, 0, 0'' means that the basic building block `A' is first combined with element 2 (explained below), then again with element 2, followed in turn by element 2, element 1, element 0, and element 0, until the full string is recovered. Next, three `@' symbols separate three elements. The first pair `@', `EF', designates element 0, meaning element 0 is created by combining the basic building blocks `E' and `F'. The second pair, `@', `CD', designates element 1, formed by combining the basic building blocks `C' and `D'. The third group, `@', `B', 1, designates element 2, which is constructed from the basic building block `B' and element 1. Note that the IDs shown in the `compressed\_list' can differ from those assigned by the Ladderpath algorithm itself, owing to implementation details.

This design is primarily motivated by operational considerations, making subsequent processing more convenient. For instance, the inclusion of explicit separators (such as `@') in the compressed list facilitates structural parsing and reconstruction. Moreover, this list can be directly used as input for entropy encoding methods--such as Huffman coding---to achieve further compression and efficient storage.

\newpage
\section{Normal compressor}
\label{App:normal}

First, the concept of $\eta$ is briefly recapped, as it will be needed later. In the Ladderpath approach, there are three primary indices for measuring a system:
\begin{itemize}
    \item Ladderpath-index ($\lambda$): The length of the shortest ladderpath for a target.
    \item Size-index ($S$): The length of the shortest trivial ladderpath for an object.
    \item Order-index ($\omega$): Defined as $\omega(x):=S(x) - \lambda(x)$, this means that the work has been saved from combining blocks when constructing the target.
\end{itemize}
Based on these definitions, the $\eta$ is calculated by the following formula:
\begin{equation*}
    \eta = \frac{\omega(x) - \omega_0(S)}{\omega_{\text{max}}(S) - \omega_0(S)}
\end{equation*}
where $\omega_0(S)$ represents the average Order-index of all possible sequences of length $S$.  $\omega_{\text{max}}(S)$ represents the maximum Order-index among all sequences of length $S$. Through the aforementioned formula, $\eta$ is used to quantify the degree of orderliness of a sequence, reflecting the structural characteristics of the system at a specific length $S$ (details can be referred to reference \textit{Zhang et al., Physical Review Research 6(2):023215, 2024}).

Then, we explain ``Normal Compressor''. A compressor that can satisfy these four properties with an error range of $\log(n)$ is called a ``Normal Compressor'' (\textit{Li et al., IEEE Transactions on Information Theory 50(12):3250–3264, 2004}).
\begin{itemize}
    \item {Idempotency:}  
        \( C(xx) = C(x) \text{~and~} C(A) = 0, \\
        \text{~when~} A = \emptyset. \)
        
    \item {Monotonicity:}  
        \( C(xy) \geq C(x). \)
        
    \item {Symmetry:}  
        \( C(xy) = C(yx). \)
        
    \item {Distributivity:}  
        \( C(xy) + C(z) \leq C(xz) + C(yz). \)
\end{itemize}

For Idempotency, with \(x =ABCDEFBCD\), \(C(x) = \lambda_x -1= 6\), \(C(xx) = \lambda_{xx}-2=6\). Combined with the reuse principle of Ladderpath, it is clear that after constructing the subsequence \(x\), it can be reused infinitely many times without consuming additional resources without consuming additional resources. In the process of combining \(xx\), we need to take out the sequence \(x\) twice, and this operation has nothing to do with the compression process, so we have \(C(xx) = \lambda_{xx}-2\).

\begin{figure}[http]
    \centering
    \includegraphics[width=0.7\columnwidth]{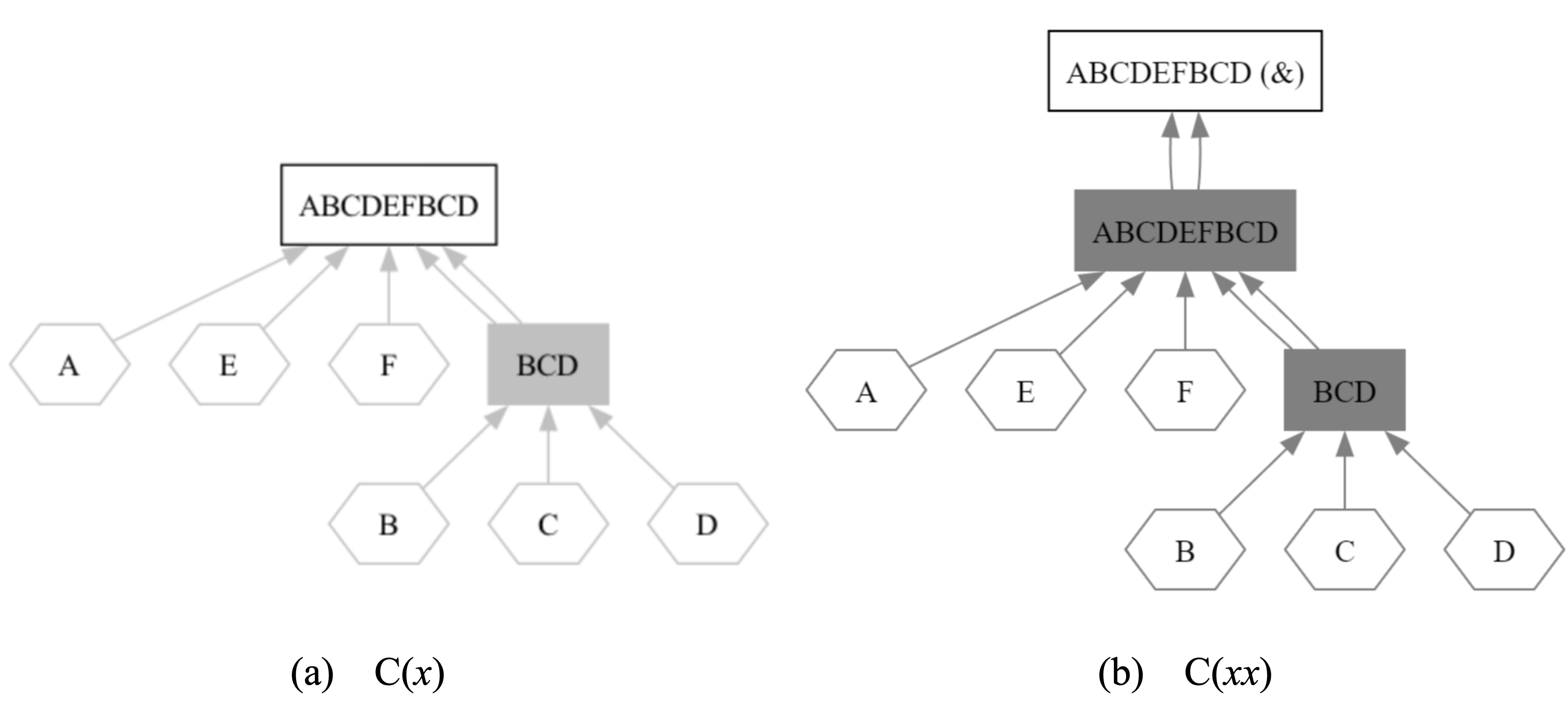}
    \caption{Laddergraph for $x$ and $xx$.}
    \label{figS2}
\end{figure}
\vspace{0.2cm}
For Monotonicity, it is clear that Idempotency is satisfied when \(y = 0\) or \(y = x\) in the presence of \(C(xy) =  C(x)\). We focus on the cases\(y \neq 0\)  as well as \(y \neq x\). We can start by going through the construction steps and reuse principles of the Ladderpath. The construction of the string $xy$ must contain the steps necessary to construct $x$; thus, its cost is at least equal to the cost of constructing $x$. Specifically, the minimum number of steps to construct $xy$: $\lambda_{xy} \geq \lambda_x$, and the savings from reuse do not affect this inequality, so we can know that: \(C(xy) \geq C(x)\). Even if $y$ introduces a new structure, its steps do not reduce the overhead of the $x$ part, thus ensuring that Monotonicity holds.

Symmetry is the same as Idempotency. The Ladderpath compressor tends to generate $x$ and $y$ first when compressing objects $xy$ and $yx$, so the difference between generating $xy$ and $yx$ is whether to take out $x$ or $y$ first, which is only the last step, and the resources for compressing $x$ and $y$ in the previous step are unchanged.

For the fourth Distributivity, there is a stronger distributive law \(C(xyz) + C(z) < C(xz) + C(yz)\), and here we continue the idea of NCD and use a slightly weaker distributive law for verification (subsequently referred to as ``weak''). Firstly, it should be noted that in the experimental design, in order to encompass as much as possible the various possibilities from purely random sequences to fully homogeneous sequences and to eliminate the influence of the number of repeated substructures within the sequence on the compression efficiency. We chose $\eta = [0, 0.1, 0.2, ..., 0.95]$, each containing 100 strings. In Table \ref{tab:eta_comparison}, we compare the different compressors that do not satisfy the distributivity for each data group under the strong and weak constraints, respectively.
\begin{table*}[!ht]
\centering
\renewcommand{\arraystretch}{1} %
\resizebox{0.9\textwidth}{!}{%
    \begin{tabular}{|c|cc|cc|cc|cc|cc|cc|}
    \hline
    \multirow{2}{*}{\textbf{$\eta$}} & \multicolumn{2}{c|}{\textbf{Ladderpath}} & \multicolumn{2}{c|}{\textbf{gzip}} & \multicolumn{2}{c|}{\textbf{LZ77}} & \multicolumn{2}{c|}{\textbf{bz2}} & \multicolumn{2}{c|}{\textbf{lzma}} & \multicolumn{2}{c|}{\textbf{Zstandard}} \\
    \cline{2-13}
     & \textbf{strong} & \textbf{weak} & \textbf{strong} & \textbf{weak} & \textbf{strong} & \textbf{weak} & \textbf{strong} & \textbf{weak} & \textbf{strong} & \textbf{weak} & \textbf{strong} & \textbf{weak} \\ \hline
     
    0    & 1   & 0   & 91  & 0   & 3   & 0   & 4   & 0   & 18  & 0   & 49  & 0   \\
    0.1  & 0   & 0   & 62  & 0   & 1   & 0   & 45  & 0   & 2   & 0   & 7   & 0   \\
    0.2  & 0   & 0   & 51  & 0   & 2   & 0   & 62  & 0   & 5   & 0   & 11  & 0   \\
    0.3  & 6   & 0   & 24  & 0   & 5   & 0   & 45  & 0   & 8   & 0   & 2   & 0   \\
    0.4  & 10  & 0   & 25  & 0   & 6   & 0   & 33  & 0   & 13  & 0   & 16  & 0   \\
    0.5  & 11  & 0   & 13  & 0   & 2   & 0   & 8   & 0   & 3   & 0   & 14  & 0   \\
    0.6  & 13  & 0   & 10  & 0   & 4   & 0   & 10  & 0   & 5   & 0   & 16  & 0   \\
    0.7  & 18  & 0   & 21  & 0   & 4   & 0   & 0   & 0   & 12  & 0   & 47  & 2   \\
    0.8  & 19  & 0   & 11  & 0   & 3   & 0   & 1   & 0   & 5   & 0   & 33  & 3   \\
    0.9  & 17  & 0   & 10  & 1   & 18  & 0   & 12  & 1   & 2   & 0   & 8   & 1   \\ \hline

    \textbf{Total} & 9.50\% & 0.00\%   & 31.80\% & 0.10\%   & 4.80\%  & 0.00\%  & 22.00\% & 0.10\%  & 7.30\%  & 0.00\%  & 20.30\% & 0.60\% \\
    \hline
    \end{tabular}%
    }
\caption{Counts of violations under different $\eta$ values for various compression methods.}
\label{tab:eta_comparison}
\end{table*}

Across different $\eta$ values, nearly all compression methods satisfy the weakly relaxed distributivity of Kolmogorov complexity within the acceptable margin of error. Ladderpath demonstrates relatively greater stability, with only two string sequences failing to meet the weak distributivity. In contrast, gzip exhibits three violations, while \textit{Zstandard} performs the worst, with eight sequences not conforming to the property. Under the strong constraints of Kolmogorov complexity, Ladderpath's anomalous values remain in an intermediate state. \textit{Gzip} shows the poorest performance, particularly when compressing disordered sequences with smaller $\eta$ values, where a substantial number of instances fail to satisfy the strong distributivity. Regardless of the chosen $\eta$-value, LZ77 consistently achieves the best performance.

\begin{figure}[http]
    \centering
    \includegraphics[width=0.6\linewidth]{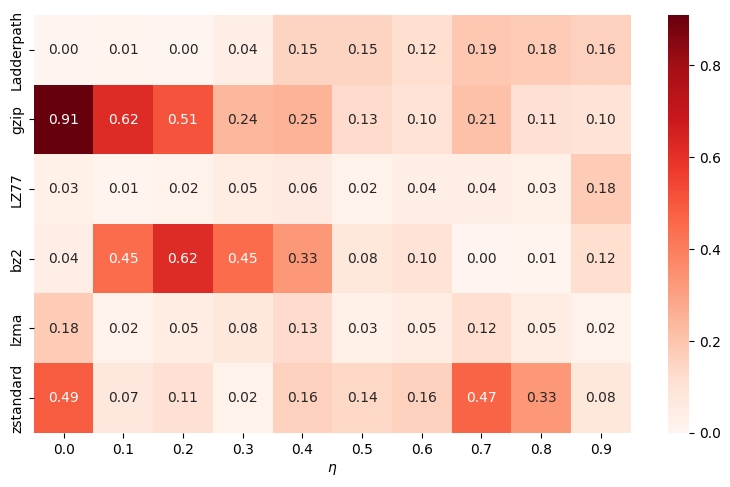}
    \caption{Compression method performance by $\eta$ value: Strong. The horizontal axis represents the orderliness parameter $\eta$ (0-1), and the vertical axis indicates different compressors. The color intensity reflects the proportion of groups that violate the strong distributive property. Darker colors (towards red) indicate poorer performance, while lighter colors (towards white) indicate better performance.}
    \label{fig:Kolmogorov_strong}
\end{figure}

To be more comprehensive, we conducted additional tests to examine the satisfaction of ``strong'' distributivity---the most challenging property to satisfy---namely, $C(xy) + C(z) \leq C(xz) + C(yz)$, as shown in Fig. \ref{fig:Kolmogorov_strong}. The results show that none of the tested compression algorithm can strictly satisfy the strong distributivity, with $gzip$ and $bzip2$ showing significant outliers and large deviations. Ladderpath-based compressor almost completely satisfies the strong distributivity for small $\eta$ (e.g., less than 0.4). It almost always satisfies the strong distributivity, while typically, natural language texts have an ordering rate $\eta$ in this range. For strong distributivity, the better the satisfaction, the more effective their discriminative distances are. In summary, the Ladderpath-based compressor shows good properties compared to some other compressors, and outperforms traditional compressors such as $gzip$ and $Zstandard$.

\section{Download datasets}
\label{App:datasets}
Detailed download links for all datasets are listed below:
\begin{itemize}

    \item AGNews: Available through \\
    \url{http://groups.di.unipi.it/~gulli/AG_corpus_of_news_articles.html}

    \item DBpedia: Accessible through \textit{TorchText} \\
    \url{https://docs.pytorch.org/text/stable/index.html}
    
    \item Reuters R8 and R52: Available from the \textit{Text Categorization Corpora} \\
    \url{https://disi.unitn.it/moschitti/corpora.htm}
    
    \item KinyarwandaNews and KirundiNews: Freely downloadable from the authors' repository \\
    \url{https://github.com/Andrews2017/KINNEWS-and-KIRNEWS-Corpus}
    
    \item SwahiliNews: Freely available via \textit{HuggingFace} \\
    \url{https://huggingface.co/datasets/community-datasets/swahili_news}
    
    \item DengueFilipino: Freely available via \textit{HuggingFace} \\
    \url{https://huggingface.co/datasets/jcblaise/dengue_filipino}
    
    \item SogouNews: Available on \textit{TorchText} \\
    \url{https://docs.pytorch.org/text/stable/datasets.html#sogounews}

    \item Code for Paper: “Low-Resource” Text Classification: A Parameter-Free Classification Method with Compressors \\
    \url{https://github.com/bazingagin/npc_gzip}
    
\end{itemize}


  
  
  
  
  
  



\section{Data preprocessing}
\label{App:preprocess}



Table \ref{datasets} presents the sizes of all datasets.
\begin{table}
\caption{Original data volume and the data volume used in our experiments.}
\label{datasets}
\begin{tabular*}{\linewidth}{@{\extracolsep{\fill}} lcccc @{}}
\toprule
\textbf{Datasets} & \textbf{RawTest} & \textbf{RawTrain} & \textbf{UsedTest} & \textbf{UsedTrain} \\
\midrule
AGNews & {7600} & {120000} & {7600 (100\%)} & {120000 (100\%)} \\
DBpedia & {70000} & {560000} & {7000 (10\%)} & {56000 (10\%)} \\
SogouNews & {60000} & {450000} & {6000 (10\%)} & {45000 (10\%)} \\
R8 & {2189} & {5485} & {2175 (cleaned)} & {5415 (cleaned)} \\
R52 & {2568} & {6532} & {2552 (cleaned)} & {6438 (cleaned)} \\
KinyarwandaNews & {4253} & {17014} & {2562 (cleaned)} & {8593 (cleaned)} \\
KirundiNews & {923} & {3689} & {671 (cleaned)} & {1705 (cleaned)} \\
SwahiliNews & {7338} & {22207} & {6951 (cleaned)} & {20912 (cleaned)} \\
DengueFilipino & {500} & {4015} & {495 (cleaned)} & {3945 (cleaned)} \\
\bottomrule
\end{tabular*}
\end{table}

\begin{figure}[http]
    \centering
    \includegraphics[width=0.5\columnwidth]{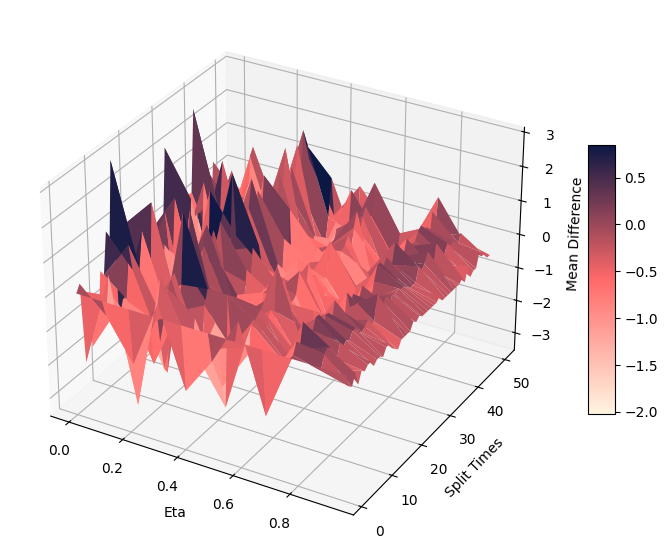}
    \caption{Impact of split strategies and $\eta$ values on sequence compression.}
    \label{fig_split}
\end{figure}

In addition, note that some datasets contain sentences of considerable length. To reduce computational cost---since the runtime of the Ladderpath approach grows with sentence length---we divided such long sentences into shorter segments. An important question, however, is whether splitting a long sentence into multiple shorter segments affects the resulting Ladderpath measure. To examine this, we conducted the following test. For each sentence, we randomly introduced from 1 to 50 cuts: one cut yields two shorter segments, two cuts yield three segments, and so on, up to 50 cuts yielding 51 segments. We then applied the Ladderpath compression independently to each truncated segment and compared the aggregate results with those obtained from the original unsplit sentence. The $z$-axis of Fig. \ref{fig_split} reports the average difference between the computed values. The parameter $\eta$ was varied over the range $[0,0.1,0.2,\ldots,0.95]$, and for each $\eta$ we analyzed 10 randomly generated strings.

Combined with the results in Fig. \ref{fig_split}, the following conclusions can be drawn:
\begin{itemize}
    \item Low $\eta$: When $\eta$ is small ($0 - 0.2$), the overall mean difference fluctuates significantly with the number of truncations. It is hypothesized that when $\eta$ is small, there are only a limited number of repeated fragments available. The truncation operation may further disrupt the redundant repetitions that could be utilized by Ladderpath, leading to a significant response in the truncated compression difference.
    
    \item Medium $\eta$: When $\eta$ takes a moderate value ($0.4-0.7$), the results indicate that most of the difference magnitude is at a medium level. This suggests that within this range of $\eta$, the repetitiveness and randomness of text sequences are relatively balanced. An increase in the number of truncations does not result in extreme variations in compression performance; in some cases, a well-designed segmentation strategy may even moderately improve compression efficiency.
    
    \item High $\eta$ (close to $1.0$): When the $\eta$ value is high, the text is more structured with a greater number of repeated subsequences. In this case, segmentation does not significantly enhance or degrade the compression performance of Ladderpath.
\end{itemize}
The above results show that for a text sequence of about 1000 lengths in this experiment, the split has a small effect on the compressed length. The average difference between splits and original sequences is mostly in the range of -2 to 0.5, which is negligible. Sequences with different $\eta$ values have different sensitivities to truncation; the larger the $\eta$ value and the more organized the sequence, the weaker the effect of truncation on the compression effect. In any case, we can see that the truncation of text does not significantly change the compression result, thus verifying the feasibility of reducing the complexity of the algorithm by cutting the sequence.


\section{Tie-breaking details in $k$-NN}
\label{App:kNNoptimistic}

$k$-NN is a widely used algorithm for classification tasks due to its intuitive approach. The fundamental idea is to classify a sample by identifying the nearest $k$ neighbors in the training set. Based on the distribution of categories among these neighbors, the sample's category is determined through voting or a weighted decision. However, challenges arise when there is a tie in the votes among the $k$ neighbors. 

There are various strategies for dealing with this problem: (1) Random: The label of a randomly selected neighbor in the flat ticket category is assigned to the sample to be classified; (2) Nearest: Among the neighbors of the tie-ballot, rank them according to their distance from the sample to be classified and assign the label of the nearest neighbor to that sample.

After resolving the tie vote, researchers usually evaluate classification accuracy by comparing predicted labels with true labels. In the context of the study conducted by Jiang et al. ($k=2$), it did not use Random or Nearest methods when facing a tie. Instead, it directly analyzes the labels of the two neighbors to provide results on classification accuracy. Their strategies for interpreting tie votes are:
\begin{itemize} 
    \item {Optimistic}: Under this strategy, the prediction is considered correct if at least one of the two neighbors' categories matches the true category during a tie scenario.
    \item {Pessimistic}: This approach demands that both neighbors must agree with the true category for the prediction to be deemed correct when a tie occurs.
\end{itemize}

Table \ref{pessimistic} and \ref{optimistic} presents the outputs from the two decision-making strategies when $k=2$, with W representing wrong classifications and C representing correct classifications. The results indicate that the Optimistic strategy achieves a correctness rate of $3/4$, while the Pessimistic strategy yields a significantly lower correctness rate of only $1/4$. Although the original article mentions a Random method, it was not practically used. This is likely due to the fact that with $k=2$, the Random choice can result in a $50\%$ error probability, which does not meaningfully enhance performance.

\begin{table}[H]
\centering
\renewcommand{\arraystretch}{1.2}
\small
\setlength\tabcolsep{8pt}     
\begin{minipage}{0.5\textwidth}
  \centering
  \begin{tabular}{ccc}
    \toprule
    \textbf{Labels} & \textbf{Voting} & \textbf{Output} \\
    \midrule
    WW & W-W & W \\
    WC & W-C & W \\
    CW & C-W & W \\
    CC & C-C & C \\
    \bottomrule
  \end{tabular}
  \caption{Pessimistic decision-making.}
  \label{pessimistic}
\end{minipage}%
\hfill
\begin{minipage}{0.5\textwidth}
  \centering
  \begin{tabular}{ccc}
    \toprule
    \textbf{Labels} & \textbf{Voting} & \textbf{Output} \\
    \midrule
    WW & W-W & W \\
    WC & W-C & C \\
    CW & C-W & C \\
    CC & C-C & C \\
    \bottomrule
  \end{tabular}
  \caption{Optimistic decision-making.}
  \label{optimistic}
\end{minipage}
\end{table}

Ultimately, the authors in the reference (\textit{Jiang et al., ACL 2023, pp. 6810–6828}) opted for the optimistic strategy, which contributed to the relatively high accuracy of their results (albeit slightly unrealistically high). In this study, we have chosen to implement the Nearest method for resolving the tie vote problem, which provides a fairer comparison.

Table \ref{tab:comparison-results_optimistic} gives the comparison results on multiple datasets using Nearest and Optimistic strategies. The difference in performance of Nearest over Optimistic with $k = 2$ is clearly visible, which is why the data in the original article is inflated.

\begin{table}[!ht]
\centering
\renewcommand{\arraystretch}{1.2} 
\scalebox{1.0}{
\begin{tabular}{lcccc}
    \toprule
    \textbf{Method} & \textbf{AGNews} & \textbf{DBpedia} & \textbf{R8} & \textbf{R52} \\
    \midrule
    Nearest & 0.790 & 0.957 & 0.890 & 0.801 \\
    Optimistic & 0.937 & 0.970 & 0.952 & 0.889 \\
    \bottomrule
\end{tabular}
}
\vspace{0.3cm}
\caption{Performance comparison of different decision-making methods under $k=2$.}
\label{tab:comparison-results_optimistic}
\end{table}

\section{Out-of-distribution case with $k = 2$}
\label{App:table2k_2}

\begin{table}[!ht]
\centering
\renewcommand{\arraystretch}{1.2} 
\scalebox{1.0}{
\begin{tabular}{lccccc}
    \toprule
    \textbf{Model} & \textbf{Kinyar-wanda-News} & \textbf{Kirundi-News} & \textbf{Dengue-Filipino} & \textbf{Swahili-News} & \textbf{SogouNews} \\
    \midrule
    BERT & \textbf{0.838} & 0.879 & 0.979 & 0.897 & 0.952 \\ 
    \midrule
    $NCD_{gzip}$ & \underline{0.835} & 0.872 & 0.956 & 0.889 & \textbf{0.957} \\
    $NCD_{lp}$ & 0.826 & 0.876 & 0.971 & 0.858 & 0.951 \\
    $L_{Dice}$ & \underline{0.835} & \textbf{0.897} & 0.976 & 0.892 & \textbf{0.957} \\
    $L_{Jaccard}$ & \underline{0.835} & \textbf{0.897} & \textbf{0.983} & \textbf{0.906} & \textbf{0.957} \\
    \bottomrule
\end{tabular}
}
\vspace{0.3cm}
\caption{Comparison of text classification accuracy across different models on five OOD datasets ($k=2$).}
\label{tab:table2k_2}
\end{table}

\newpage
\section{Out-of-distribution case in few-shot setting with $k = 2$}
\label{App:table3k_2}

\begin{table}[!ht]
\centering
\renewcommand{\arraystretch}{1.2} 
\scalebox{1.0}{
\begin{tabular}{lccccc}
    \toprule
    \textbf{Model} & \textbf{Kinyar-wanda-News} & \textbf{Kirundi-News} & \textbf{Dengue-Filipino} & \textbf{Swahili-News} & \textbf{SogouNews} \\
    \midrule 
    BERT & 0.240 & 0.386 & 0.409 & 0.396 & 0.221 \\ 
    \midrule 
    $NCD_{gzip}$ & 0.280 & \textbf{0.456} & 0.324 & \textbf{0.551} & 0.508 \\
    $NCD_{lp}$ & 0.249 & 0.398 & 0.376 & 0.522 & 0.502 \\
    $L_{Dice}$ & \textbf{0.285} & \textbf{0.456} & \textbf{0.416} & \textbf{0.551} & \textbf{0.583} \\
    $L_{Jaccard}$ & \textbf{0.285} & \textbf{0.456} & \textbf{0.416} & \textbf{0.551} & \textbf{0.583} \\
    \bottomrule 
\end{tabular}
}
\vspace{0.3cm}
\caption{Comparison of text classification accuracy across various methods on OOD datasets in the 5-shot setting using different methods with $k=2$.}
\label{tab:table3k_2}
\end{table}

\end{document}